\RequirePackage{fix-cm}
\documentclass[smallextended]{svjour3}       
\smartqed  
\usepackage{amsmath}
\usepackage{amssymb}
\usepackage{graphicx}
\usepackage{color}
\usepackage{subcaption}
\begin{document}

\title{Predicting How to Distribute Work Between Algorithms and Humans to Segment an Image Batch}

\subtitle{}

\titlerunning{Distributing Work Between Algorithms and Humans}        

\author{Danna Gurari \and Yinan Zhao \and Suyog Dutt Jain \and Margrit Betke \and Kristen Grauman}


\institute{Danna Gurari and Yinan Zhao and Suyog Dutt Jain and Kristen Grauman\at
              The University of Texas at Austin\\
              2317 Speedway, Stop D9500, Austin, TX 78712\\
              \email{\{dgurari,suyog,yinanzhao,grauman\}@utexas.edu}
           \and
           Kristen Grauman \at 
           Facebook AI Research 
           \and
           Suyog Jain \at 
           CognitiveScale
           \and
			Margrit Betke\at
              Boston University\\
              111 Cummington Mall, Boston, MA 02215\\
              \email{\{betke\}@cs.bu.edu}
              }

\date{Received: date / Accepted: date}

\maketitle

\begin{abstract}
Foreground object segmentation is a critical step for many image analysis tasks.  While automated methods can produce high-quality results, their failures disappoint users in need of practical solutions.  We propose a resource allocation framework for predicting how best to allocate a fixed budget of human annotation effort in order to collect higher quality segmentations for a given batch of images and automated methods.  The framework is based on a prediction module that estimates the quality of given algorithm-drawn segmentations.  We demonstrate the value of the framework for two novel tasks related to predicting how to distribute annotation efforts between algorithms and humans.  Specifically, we develop two systems that automatically decide, for a batch of images, when to recruit humans versus computers to create 1) coarse segmentations required to initialize segmentation tools and 2) final, fine-grained segmentations.  Experiments demonstrate the advantage of relying on a mix of human and computer efforts over relying on either resource alone for segmenting objects in images coming from three diverse modalities (visible, phase contrast microscopy, and fluorescence microscopy).
\end{abstract}


\section{Introduction}
A common question people ask when needing to annotate their images is whether automated options are sufficient or they should instead bring humans in the loop to create accurate annotations.  We explore this question for the task of demarcating object regions, i.e., creating \textit{foreground object segmentations}.  Foreground object segmentation is important for many downstream tasks including collecting measurements (features), differentiating between types of objects (classification), and finding similar images in a database (image retrieval).  Our goal is to intelligently distribute segmentation work between humans and computers when human effort is limited.    

\begin{figure}[t!]
\centering
\begin{subfigure}{1\textwidth}
  \centering
  \includegraphics[width=1\linewidth]{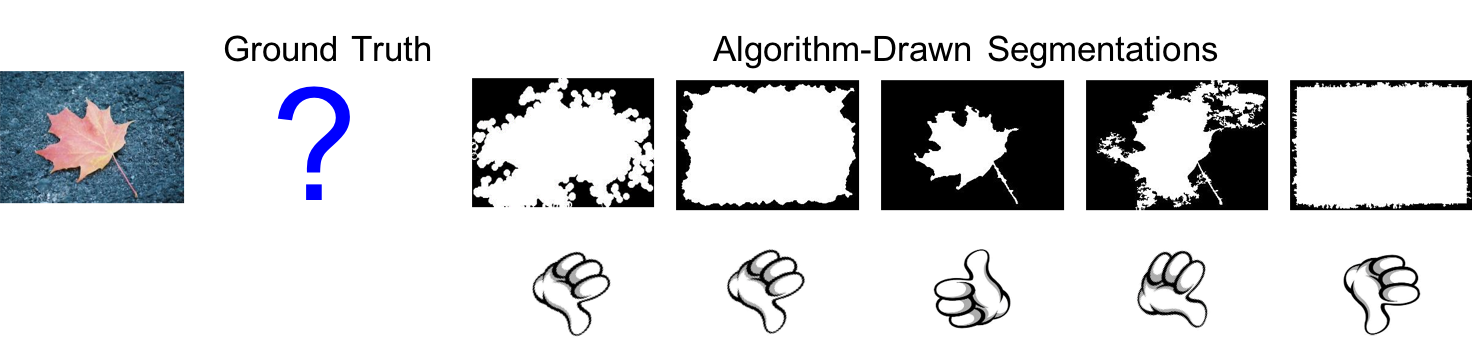}
    \caption{}
    \vspace{2em}
\end{subfigure}
\begin{subfigure}{0.6\textwidth}
  \centering
  \includegraphics[width=1\linewidth]{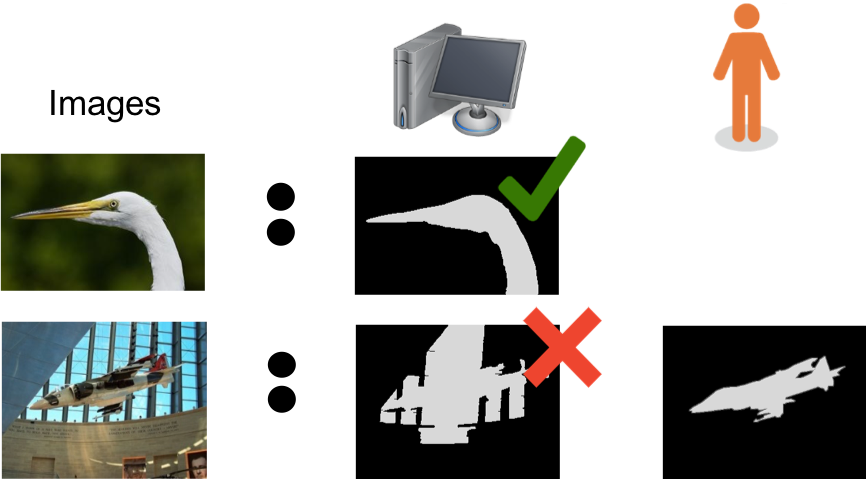}
  \caption{}
\end{subfigure}
\caption{We propose a task of predicting the quality of an image segmentation compared to the unseen ground truth in order to automatically (\textbf{a}) predict which among multiple algorithms will yield the highest quality segmentation and then (\textbf{b}) decide when to ``pull the plug" on computers and use humans instead to create high quality segmentations.}
\label{fig_motivationFigure}
\end{figure}

Our work is partially inspired by the observation that fully-automated algorithms can produce high-quality foreground object segmentations when they are successful, yet their performance often is inconsistent on diverse datasets.  This is because algorithms embed assumptions about how to separate an object from the background that are relevant for particular types of images, yet restrict their widespread applicability~\cite{Ballard81,ChanVe01,LanktonTa08,Otsu79,RotherKoBl04}.  Consequently, the knowledge of when segmentation algorithms will succeed is currently a highly-specialized skill often resigned to computer vision experts or applications specialists who spent years studying the algorithms.  Moreover, many researchers agree that there is not a one-size-fits-all segmentation solution.  Thus, lay persons needing \textit{consistently} high quality segmentations currently face a brute force approach of reviewing all images with available algorithm-drawn segmentations to decide which algorithm is best-suited per image (\textbf{Figure~\ref{fig_motivationFigure}a}) and when to enlist human effort to re-annotate images because the best-suited algorithm produces a poor quality result (\textbf{Figure~\ref{fig_motivationFigure}b}).

Our work is also inspired by the observation that widely-used segmentation tools that rely on \textit{initialization} are often inefficient because of their exclusive reliance on human input~\cite{CarlierChSaGiMa14,GradyJoSe11,GurariThSaBe14,JainGr13,LempitskyKoRoSh09,RotherKoBl04,WuZhZhLuTu14}. Specifically, humans create initial bounding boxes or coarse segmentations to localize the object of interest in every image.  A motivation for leveraging human guidance per image is that a segmentation tool can only succeed when initializations are sufficiently close to the true object boundary~\cite{JainGr13}.  A weakness of relying on humans is that for numerous methods, including level set based methods~\cite{BernardFrThUn09,ChanVe01,LanktonTa08,LiKaGoDi08}, users typically have to wait for minutes or more per image to validate whether the tool successfully converts their coarse input into high quality segmentations.  Intuitively, one may expect that computers at times can create good enough segmentations to replace human initialization effort and so minimize human effort both for initialization and validation of the results.  Still, lay persons typically lack the expertise to decide which images to distribute to computers.

We propose techniques to predict how to distribute annotation efforts between algorithms and humans for segmenting images.  We address two novel tasks.  First, we propose a system that intelligently allocates computer effort to replace human effort in order to create initial coarse object segmentations for refinement by segmentation tools.  Second, we propose a system that automatically identifies images to have humans re-annotate from scratch by predicting which images the refinement methods segmented poorly.  With both systems, a user provides a batch of images and indicates his/her available time for image annotation.  In return, the system automatically decides for each image which algorithm will yield the best results and guides the user to only annotate images deemed to be most difficult for the available algorithms.  More broadly, our systems could be exploited to efficiently create segmentations as input for downstream tasks, such as object recognition and tracking.  We publicly share our code to support reproducing this work and future extensions ({\tt http://vision.cs.utexas.edu/HybridAlgorithmCrowdSystems/PullThePlug}).  

\section{Related Work}
Interactive \emph{co-segmentation} methods address the issue of relying on human input to initialize segmentation tools for every image in a batch~\cite{BatraKoPaLuCh10,CuiYaWeWuaZhGoTa08,LiMeLuZh14}.  However, unlike our approach, these methods require that all images in the batch show related content (e.g., dogs).  Moreover, interactive co-segmentation involves continual back-and-forth with an annotator to incrementally refine the segmentation.  Avoiding a continual back-and-forth is particularly important for segmentation tools such as level set methods~\cite{ChanVe01,LanktonTa08} that take on the order of minutes or more per image to compute a segmentation from the initialization.  We instead recruit human input at most once per image and consider the more general problem of annotating unrelated, unknown objects in a batch.

Our aim to minimize human involvement while collecting accurate image annotations is shared by active learning~\cite{Settles10}.  Specifically, active learners try to identify the most impactful, yet least expensive information necessary to train accurate prediction models~\cite{BiswasPa13,Settles10,VijayanarasimhanGr11}.  For example, some methods iteratively supplement a training dataset with images predicted to require little human annotation time to label~\cite{VijayanarasimhanGr11}.  Other methods actively solicit human feedback to identify features with stronger predictive power than those currently available~\cite{BiswasPa13}.  Unlike active learners, which leverage human input at \textit{training-time} to improve the utility of a single algorithm, our method leverages human effort at \textit{test-time} to recover from failures by different algorithms.  

Our novel tasks rely on a module to estimate the quality of computer-generated segmentations.  Related methods find top ``object-like" region proposals for a given image~\cite{ArbelaezPoBaMaMa14,CarreiraSm10,EndresHo10,JainXiGr17,KohlbergerSiAlBaGr12}.  However, most of these methods are inadequate for ranking ``object-like" proposals across a batch of images because they only return relative rankings of proposals per image~\cite{EndresHo10}.  Another method proposes an absolute segmentation difficulty measure based on the image content alone~\cite{LiuXiPuSh11}.  However, this method does not account for the different performances that are observed from different segmentation tools when applied to the same image.

Our prediction framework most closely aligns with methods that predict the error/quality of a given algorithm-drawn segmentation in absolute terms~\cite{CarreiraSm10,KohlbergerSiAlBaGr12}.  In particular, we also perform supervised learning to train a regression model.  However, prior work trained prediction models using segmentations created by a single popular algorithm (coming from the medical~\cite{KohlbergerSiAlBaGr12} and computer vision~\cite{CarreiraSm10} communities respectively).  In contrast, our model is trained using a diversity of popular algorithms from different communities applied to images coming from three imaging modalities (visible, phase contrast microscopy, fluorescence microscopy).  Specifically, we populate our training data with 14 algorithm-generated segmentation algorithms per image as well as ground truth data to capture a rich diversity of the possible quality of segmentations.  Our approach consistently predicts well, outperforming a widely-used method~\cite{CarreiraSm10} on four diverse datasets.  Our experiments demonstrate the value of our prediction model for intelligently deciding which among multiple segmentation algorithms is preferable for each image.

More broadly, our work is a contribution to the emerging research field at the intersection of human computation and computer vision to build hybrid systems that take advantage of the strengths of humans and computers together.  For example, hybrid systems combine non-expert and algorithm strengths to perform the challenging fine-grained bird classification task typically performed by experts~\cite{BransonGrWaPeBe14,WahMaBe15}.  Another system decides how much human effort to allocate per image in order to segment the diversity of plausible foreground objects in a batch of images~\cite{GurariHeXiZhSaJaScBeGr18}.  While our hybrid system design demonstrates the advantages of combining human and computer efforts, our work differs by deciding how to distribute work between more costly crowd workers and less expensive algorithms for the image segmentation task.

We initially presented these ideas at the IEEE Conference on Computer Vision and Pattern Recognition (CVPR) 2016~\cite{GurariJaBeGr16}.  This work offers considerable redesigns to all methods which in turn yields significant improvements in our experimental results.  Specifically, we propose an improved approach for predicting the quality of an algorithm-drawn segmentation by employing a larger training dataset (created using a larger collection of candidate algorithms) with an expanded feature set and ensemble regression model.  We also introduce a hierarchical, two-stage prediction system that predicts which is the best algorithm per image to produce the initialization fed to the refinement-algorithm and then predicts the quality of the output from the refinement-algorithm in order to decide whether to solicit human input.  Experimental results reveal our redesigned methods yield significant improvements for predicting the segmentation quality and producing high quality segmentations.  We also expanded our experiments to explore the performance of our models and systems when using different feature sets and testing with different datasets in order to learn when, how, and why they succeed versus fail.  

\section{Segmentations by Humans or Computers?}  
\label{sec_initByHumanOrComputer}
We first describe two prediction systems for creating different levels of segmentation detail (\textbf{Section~\ref{sec_humanVsComputerSystem}}).  Then, we describe the module used by both systems to predict the quality of algorithm-generated segmentations (\textbf{Section~\ref{sec_predictingSegQuality}}).  

\subsection{Batch Allocation of Humans \& Computers} 
\label{sec_humanVsComputerSystem} 
\begin{figure}[t!]
\centering
  \includegraphics[width=0.98\linewidth]{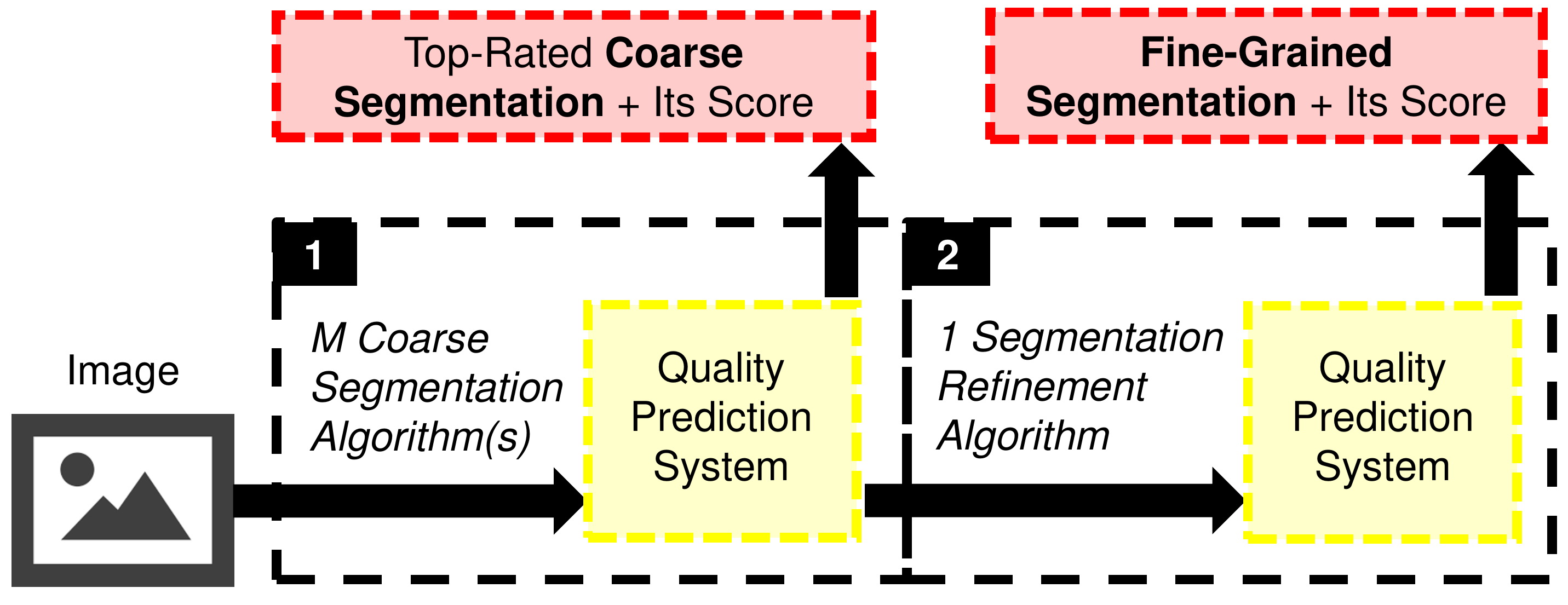}
  \caption{Overview of the relationships between the coarse segmentation system, fine-grained segmentation system, and quality prediction module.  Given an image, the coarse segmentation system applies multiple segmentation algorithms and outputs the top-predicted algorithm-generated result with its quality score.  Given the coarse segmentation output by the coarse-segmentation system, the fine-grained segmentation system applies a refinement algorithm to it and outputs the resulting segmentation with its predicted quality score. 
}
  \label{fig_coarseVsFineGrained}
\end{figure}

Our resource allocation framework predicts for each image in a batch whether the annotation should come from a human or computer.  We call this framework \emph{PTP} to reflect that the system predicts whether to ``\textbf{P}ull \textbf{T}he \textbf{P}lug" on computers and solicit human effort for each image.  We implement two \emph{PTP} systems with the goals of creating coarse and fine-grained foreground object segmentations respectively.  We examine the value of our systems with segmentation tools that require initialization.  These tools are well-suited for studying both systems because they require coarse object segmentation input and aim to output high quality, fine-grained object segmentations.  \textbf{Figure~\ref{fig_coarseVsFineGrained}} summarizes the relationship between the coarse segmentation system, fine-grained segmentation system, and the segmentation quality prediction method.  

Like existing interactive segmentation methods, we assume the user is interested in a primary foreground object~\cite{CarlierChSaGiMa14,GradyJoSe11,LempitskyKoRoSh09,RotherKoBl04,WuZhZhLuTu14}.  That is, there is a primary object of interest that the user wishes to isolate from the background.  Foreground object segmentation is therefore distinct from natural scene segmentation, where methods aim to segment all objects present in the image or delineate their boundaries or primary contours~\cite{ArbelaezMaFoMa11,EveringhamGoWiWiZi10,MartinFoTaMa01}.

 \begin{figure}[t!]
\centering
\begin{subfigure}{1\textwidth}
  \centering
  \includegraphics[width=0.80\linewidth]{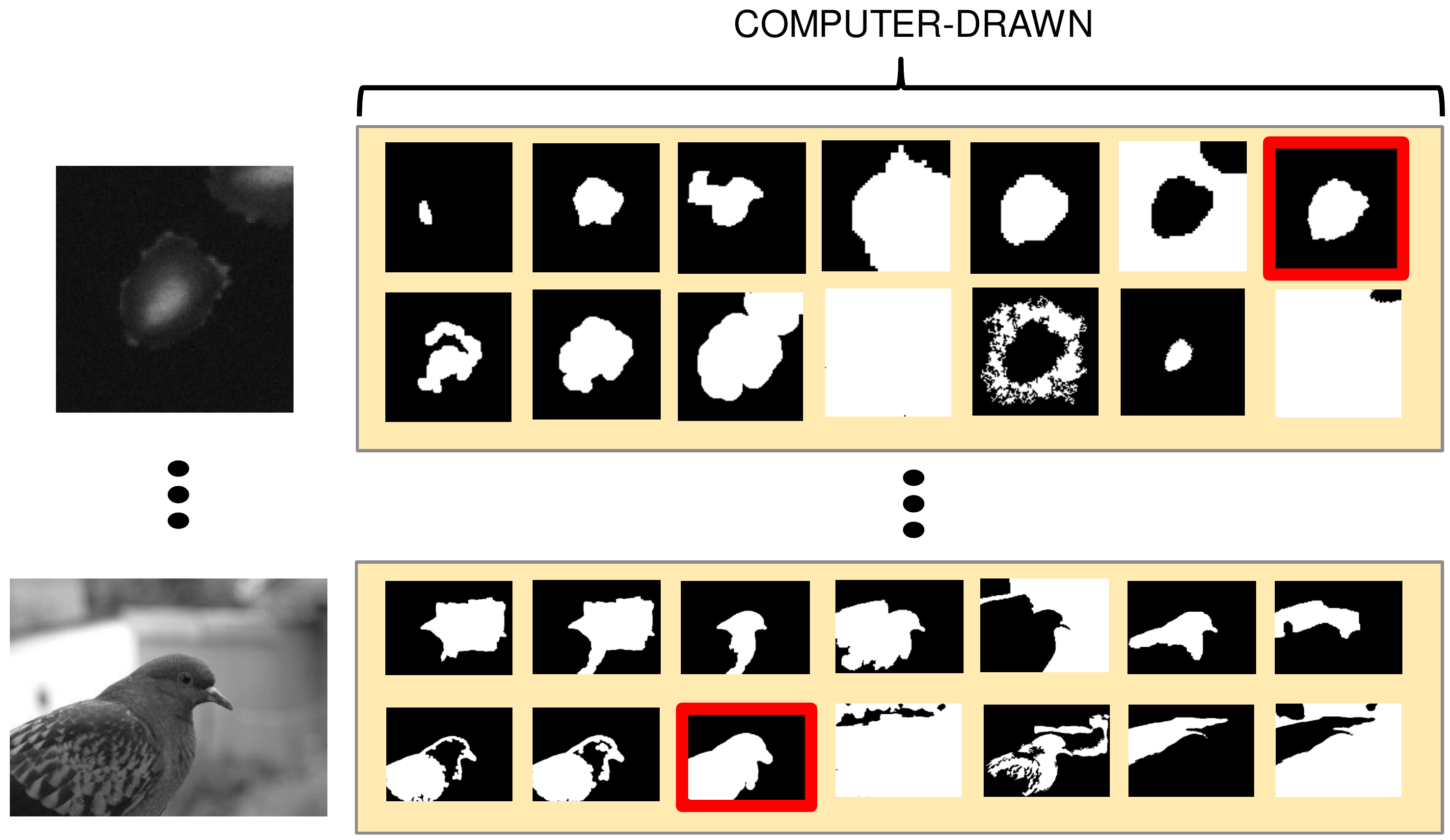}
    \caption{}
    \vspace{2em}
\end{subfigure}
\begin{subfigure}{0.90\textwidth}
  \centering
  \includegraphics[width=0.88\linewidth]{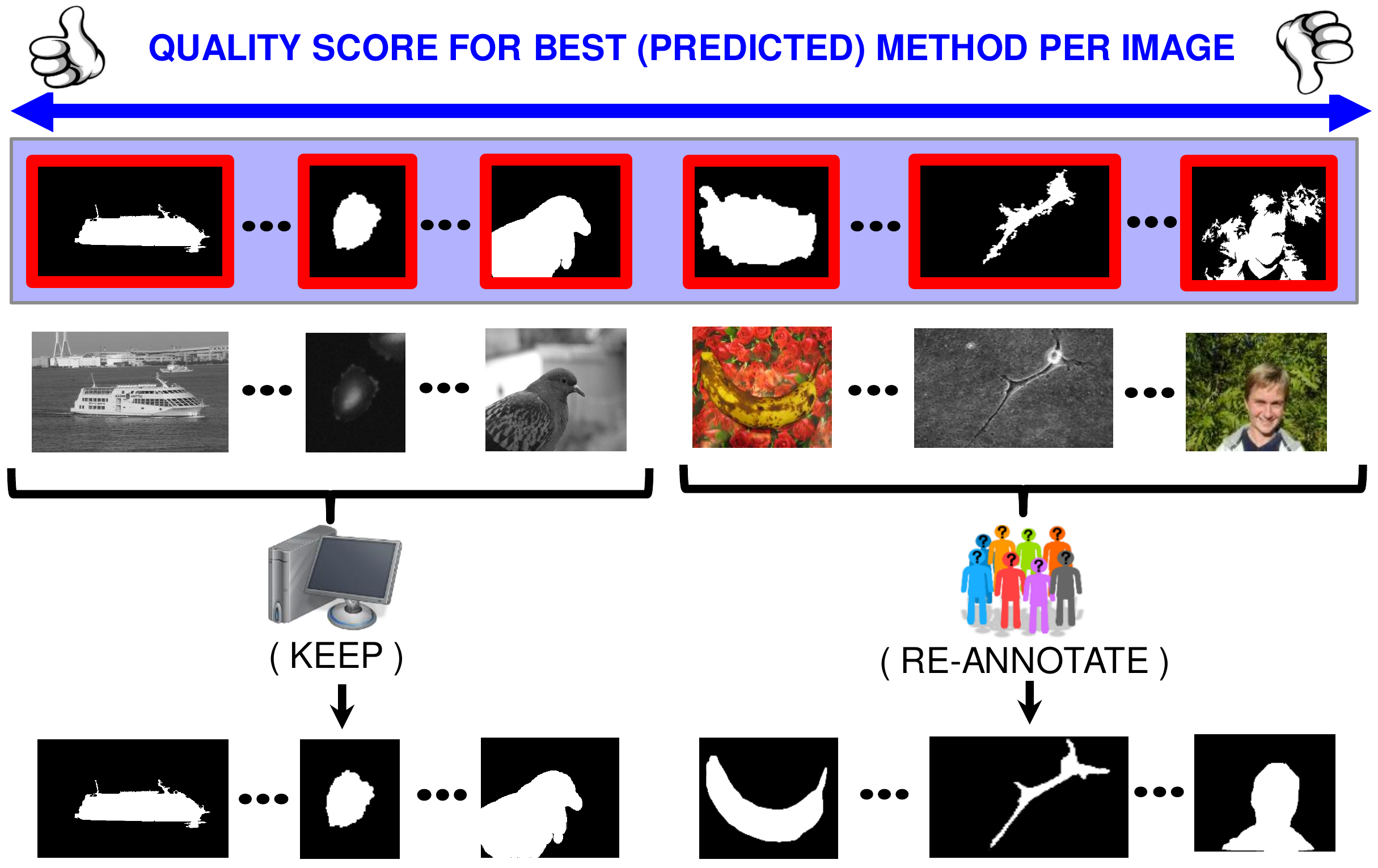}
  \caption{}
\end{subfigure}
\caption{Overview of the segmentation initialization system.  Given a batch of images, (\textbf{a}) the system automatically pairs each image with the resulting segmentation from multiple algorithms that is predicted to be the highest quality (highlighted in red).  Shown are the 14 options per image in the following order: top-3 MCG proposals~\cite{ArbelaezPoBaMaMa14}, top-3 CPMC proposals~\cite{CarreiraSm10}, salient object segmentation~\cite{LiuYuSuWaZhTaSh11}, Hough Transform with circles from radii with 3, 5, and 10~\cite{Ballard81}, adaptive thresholding and its complement, and finally Otsu thresholding and its complement~\cite{Otsu79}.  (\textbf{b}) Then, the system produces a relative ordering of all images based on the predicted quality of all selected best computer-generated results.  Finally, the system automatically allocates the available human annotation budget to images with the predicted lowest quality segmentations and keeps the automated results for the remainder of the images.}
\label{fig_ptpSystem}
\end{figure}

\subsubsection{Coarse Segmentation: Computer or Human?} 
\label{sec_coarseSegSystem}
Our first system automatically decides when to delegate the task of creating \textit{coarse segmentations} refined by segmentation tools to computers in an effort to improve upon today's status quo of relying exclusively on human input~\cite{BatraKoPaLuCh10,CuiYaWeWuaZhGoTa08,LiMeLuZh14}.  We intentionally designed the system to be agnostic to the particular refinement segmentation tool.  We implemented the system to run a refinement segmentation tool exactly once per image with one input since some tools are time-consuming to run~\cite{ChanVe01,LanktonTa08}, requiring minutes or more to refine a single initialization.  In the interest of increasing the chance of computer success, our system predicts which from a larger list of 14 computer-generated results is best-suited to create the coarse segmentation input per image.  Then, our system decides for each image whether to deploy the top-rated computer-generated coarse segmentation versus instead enlist a human to produce the initial coarse segmentation.

\textbf{Figure~\ref{fig_ptpSystem}} exemplifies the six steps of our initialization system.  First, the system collects 14 algorithm-drawn foreground segmentations per image described in Section~\ref{sec_predictingSegQuality}), then predicts the quality of each candidate segmentation using our proposed prediction system discussed in Section~\ref{sec_predictingSegQuality}, and then deploys the top-scoring option as the computer choice (\textbf{Figure~\ref{fig_ptpSystem}a}).  Next, all images are sorted based on the selected computer choices, from highest to lowest predicted quality scores, and then the system allocates the available human budget to create coarse segmentations for the allotted number of images with the lowest predicted quality scores (\textbf{Figure~\ref{fig_ptpSystem}b}).  In other words, the system relies on human effort only for the images where the computer is predicted to have the worst chance to create accurate coarse segmentations.  Finally, the system feeds all coarse segmentations created by humans and computers to the segmentation tool of interest for refinement.  

For the candidate algorithms chosen to produce computer-drawn coarse segmentations, we were motivated to employ fully-automated methods that consistently yield high-quality segmentations across the various image modalities investigated in this paper (visible, phase contrast microscopy, fluorescence microscopy).  Towards this aim, we rely on 14 variants of six algorithms discussed in current literature for foreground object segmentation both for the mainstream computer vision community~\cite{ArbelaezPoBaMaMa14,CarreiraSm10,LiuYuSuWaZhTaSh11} as well as the biomedical imaging community \cite{ChittajalluFlKoIwOrWeDaMi15,GlennLePaWeYaLuLeWaCo15,MaitraGuMu12}.  Specifically, included are the top-ranked segmentations output by two region proposal methods: multiscale combinatorial grouping (i.e., MCG)~\cite{ArbelaezPoBaMaMa14}, and constrained parametric min-cuts (i.e., CPMC)~\cite{CarreiraSm10}.  Also included is a salient object segmentation method which establishes a segmentation using a combination of local, regional, and global statistics for an image~\cite{LiuYuSuWaZhTaSh11}.  Finally, our system produces segmentations using three popular biomedical image algorithms~\cite{ChittajalluFlKoIwOrWeDaMi15,GlennLePaWeYaLuLeWaCo15,MaitraGuMu12}: Hough Transform with Circles~\cite{Ballard81}, Otsu Thresholding~\cite{Otsu79}, and adaptive thresholding.  We increase the number of options by employing the following variants of the aforementioned methods; i.e., using the top three region proposals per method~\cite{ArbelaezPoBaMaMa14,LiuYuSuWaZhTaSh11}, augmenting the image complement of the segmentation for both thresholding methods~\cite{Otsu79}, and employing different radius values for the algorithm~\cite{Ballard81} (i.e., 3, 5, 10).  Our system then post-processes each binary mask by filling all holes and keeping only the largest object.    

\subsubsection{Fine-Grained Segmentation: Computer or Human?} 
\label{sec_fineGrainedSystem}
A related yet more challenging task is predicting whether a computer-generated segmentation captures the fine-grained details describing a true object region or whether humans should instead segment images from scratch.  Whereas the first system above elicits coarse human input to initialize a segmentation tool, we next propose a second system that elicits fine-grained human input to replace segmentation tools when they segment images poorly.  The motivation of the system design is to offer a better solution than today's status quo of humans reviewing all images with associated segmentations to spot algorithm failures.   

This system consists of five key steps to segment a given batch of images.  First, a coarse segmentation is automatically generated for every image using the aforementioned \emph{Coarse Segmentation} system to choose the best computer-drawn segmentation per image from 14 options (see Figure~\ref{fig_coarseVsFineGrained}, part 1).  Then, each coarse segmentation is refined by a segmentation tool to produce a final segmentation for each image.  Next, the prediction system discussed in Section~\ref{sec_predictingSegQuality} is applied again to estimate the quality of each resulting segmentation (see Figure~\ref{fig_coarseVsFineGrained}, part 2).  Then, the system sorts all images from highest to lowest predicted quality scores for the resulting segmentations.  Finally, the system allocates the available human budget to create fine-grained segmentations for the allotted number of images with the lowest predicted quality scores.  

\subsection{Predicting Segmentation Quality}
\label{sec_predictingSegQuality}
\begin{figure}[h]
\centering
\begin{subfigure}{0.65\textwidth}
  \centering
  \includegraphics[width=1\linewidth]{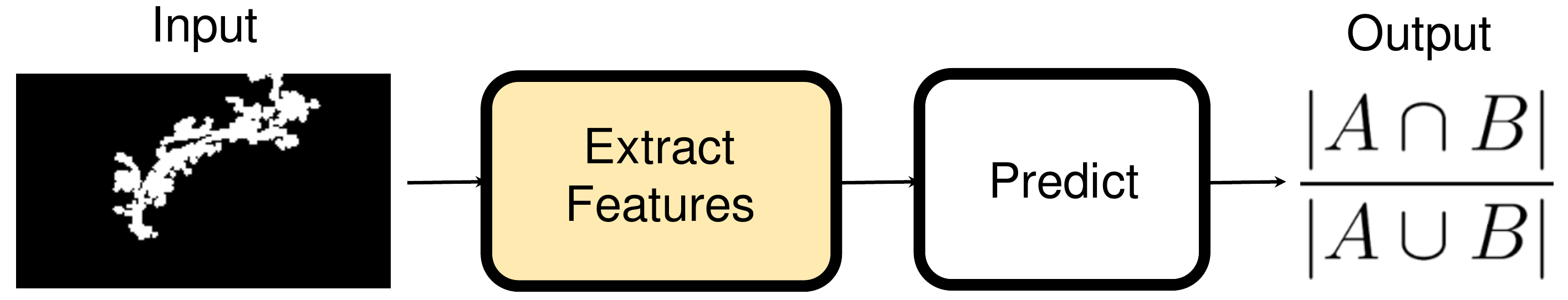}
    \caption{}
    \vspace{2em}
\end{subfigure}
\begin{subfigure}{1\textwidth}
  \centering
  \includegraphics[width=1\linewidth]{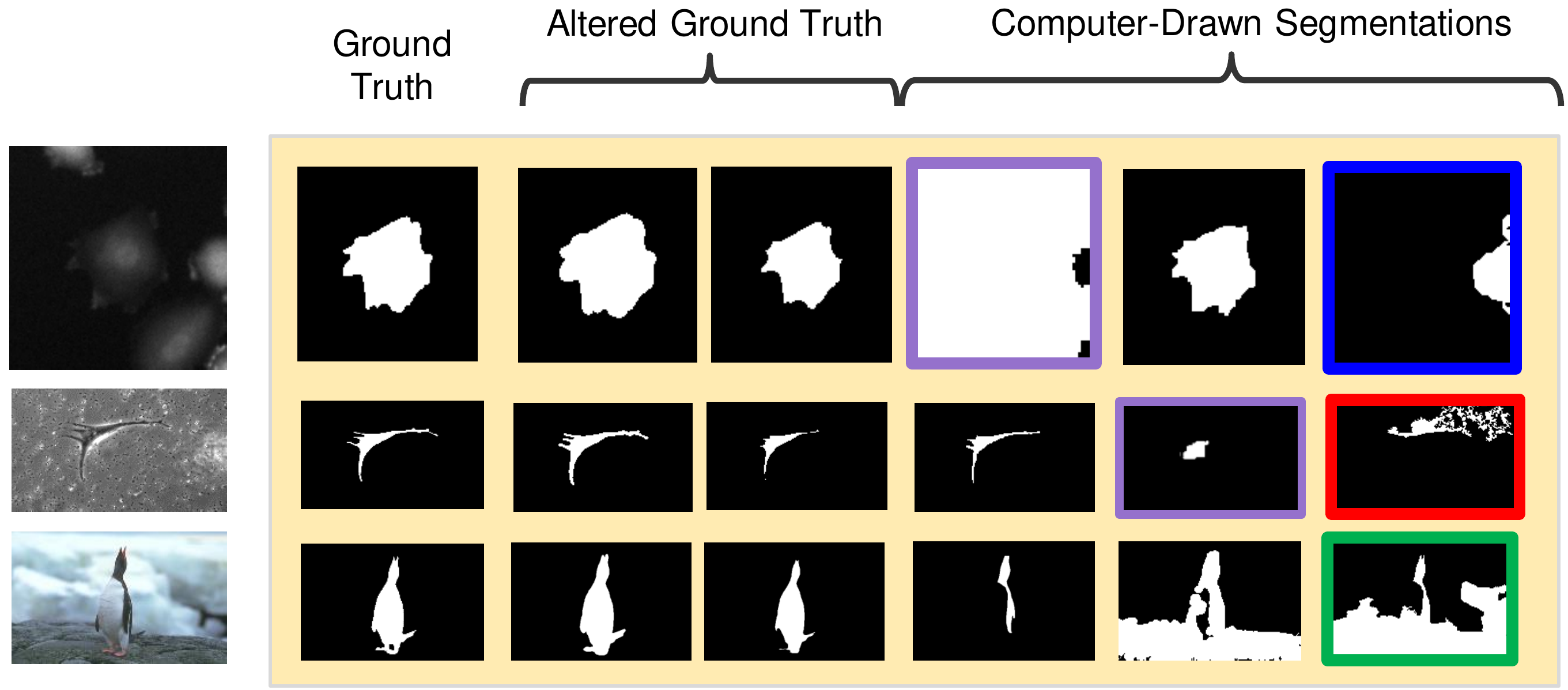}
  \caption{}
\end{subfigure}
\caption{(\textbf{a}) At test time, given a segmentation of an image, our system predicts a score indicating its similarity to the unobserved ground truth.  (\textbf{b}) To train our prediction system, we employ images showing foreground objects from three diverse domains: fluorescence microscopy (row 1), phase contrast microscopy (row 2), and visible spectrum imaging (row 3).  Examples of algorithm-generated results illustrate our training data captures a wide range of segmentation outcomes spanning from perfect (i.e., ground truth) to various failure behaviors (i.e., from different algorithms).  As shown, cues of algorithm failures are observed in the segmentation's boundary (highlighted in red), compactness (highlighted in green), location (highlighted in blue) and its image coverage (highlighted in purple).  }
\label{fig_trainingDataExamples}
\end{figure}

Embedded in both the \textit{Coarse} and \textit{Fine-Grained} segmentation systems above is a module which automatically predicts the similarity of a given segmentation to an unseen ground truth segmentation (\textbf{Figure~\ref{fig_trainingDataExamples}a}).  We propose as our prediction framework a regression model in order to capture that algorithms can produce segmentations that range in quality from complete failures to nearly perfect.  

\subsubsection{Training Instances}  
We aim to populate our training data with segmentation masks that reflect a large, relatively balanced number of examples for each segmentation quality from the range of possible segmentation qualities.  Towards this aim, we choose segmentation masks that capture the transition of segmentation quality from perfect (i.e., ground truth), to reasonable human mistakes (i.e., manipulated ground truth), to a variety of possible failure behaviors (i.e., various algorithms).  Accordingly, for each image, our system produces multiple training examples derived from the human-drawn ground truth as well as 14 algorithm-drawn segmentations.

For algorithm-drawn segmentation masks, we employ the same 14 methods used in the \emph{Coarse Segmentation} system described in the Section~\ref{sec_humanVsComputerSystem}, which includes region proposals~\cite{ArbelaezPoBaMaMa14,CarreiraSm10}, salient object segmentation~\cite{LiuYuSuWaZhTaSh11}, and popular biomedical image segmentation algorithms~\cite{ChittajalluFlKoIwOrWeDaMi15,GlennLePaWeYaLuLeWaCo15,MaitraGuMu12}.  The variety of failure behaviors produced by the different algorithms are exemplified in \textbf{Figure~\ref{fig_trainingDataExamples}b}, columns 4--6. 

Given that the training data may be insufficiently populated with higher-scoring segmentations (if all algorithms consistently fail), our system augments three binary masks based on the ground truth segmentations.  The system uses the ground truth directly in order to capture during training the appearance of a perfect segmentation.  Our system also dilates and erodes the ground truth binary mask by three pixels to simulate a slightly under-segmented and over-segmented segmentation respectively where fine details may get smoothed out or chopped off (e.g., \textbf{Figure~\ref{fig_trainingDataExamples}b; columns 3--4}).

\subsubsection{Training Data - Features}
\label{sec_trainingFeatures}
Next, our motivation is to use knowledge about algorithm behavior on everyday and biomedical images to choose predictive features.  We take advantage of the observation that the chosen algorithms sometimes fail big when they fail, manifesting appearances unlike what one would expect from widely meaningful object shapes (\textbf{Figure~\ref{fig_trainingDataExamples}b}).  We propose nine features that describe the binary segmentation mask to capture these failure behaviors.  We also consider image descriptors based on convolutional neural networks (i.e., CNNs).  We hypothesize that, in aggregation, these features may account for objects of different shapes, sizes, and appearances.  We describe these features below.

\textit{Segmentation Boundary.}  When algorithms fail, resulting segmentations often have boundaries characterized by an abnormally large proportion of highly-jagged edges.  We implement two boundary-based features to capture this observation.  We compute the \textit{mean} and \textit{standard deviation of the Euclidean distance of every point on the segmentation boundary to the centroid}.  The boundary is defined as all pixels on the exterior of the object in a binary mask using an 8-connected neighborhood.  The centroid is defined as the center of mass of the segmentation in the binary mask.  

\textit{Segmentation Compactness.} When algorithms fail, segmentations often are not compact.  We designate three features to capture this observation.  Two measures compute the coverage of segmentation pixels within a bounding region.  \textit{Extent} is defined as the ratio of the number of pixels in the segmentation to the number of pixels in the area of the bounding box.  \textit{Solidity} is defined as the ratio of the number of pixels in the segmentation to the number of pixels in the area of the convex hull.  We also compute the \textit{shape factor} to capture the circularity of the segmentation since a pure circle is a good measure to indicate highly compact objects.  It is defined as the ratio of region area $A$ to a circle with the same perimeter $P$:  $\frac{4 \pi A}{P^2}$. 

\textit{Location of Segmentation in Image.} When algorithms fail, resulting segmentation regions often lie closer to the edges of images.  This observation stems in part from the center bias of many existing datasets. We compute the \textit{normalized x} and \textit{y centroid coordinates} of the segmentation centroid in the image to capture this observation.  Specifically, we compute the x value of the center of mass divided by the image width and y value of the center of mass divided by the image height.

\textit{Coverage of Segmentation in Image.}  When algorithms fail, resulting segmentations often cover abnormally large and small areas in the image.  We implement two features to capture this observation.  First, we compute the \textit{fraction of pixels in the image that belong to the segmentation}.  Second, we compute the \textit{fraction of pixels in the image that belong to the bounding box of the segmentation}. 

\textit{Image-based CNN Features.} The above features capture elements likely to be informative for the task, based on domain knowledge of the binary segmentation mask.  As a counterpoint, we also consider feature vectors extracted from three off-the-shelf CNN architectures in order to describe the intensity values of the segmentation mask.  Specifically, we use CNN features coming from three classification systems which were pre-trained on ImageNet: AlexNet~\cite{KrizhevskySuHi12}, VGG~\cite{SimonyanZi14}, and ResNet~\cite{HeZhReSu16}.  For AlexNet, we use the last fully connected layer to create a 4096-dimensional vector.  For VGG, we use the fc7 layer to create a 4096-dimensional vector.  For ResNet, we use the pool5 layer after global average pooling to obtain a 2048-dimensional vector.
We compute these feature representations using the image patch created by using the bounding box of the segmentation.  

See Section~\ref{sec_ExperimentsAndResults} for an analysis of the variability of several of the mask-based cues measured for objects observed within diverse datasets.  This analysis highlights the variability and biases available in a range of unrelated datasets.

\subsubsection{Training Data - Labels and Regression Model} 
To create each output label, the system computes a score indicating the quality of each training instance segmentation.  We use the Jaccard index (i.e., intersection over union, IOU) which indicates the fraction of pixels that are in common to the training instance and ground truth segmentation (i.e., $\frac{|A \cap G|}{|A \cup G|}$).  

For our model, we train a regression tree ensemble with the aforementioned training data to predict the quality of a given segmentation of an image.  This model is trained to learn the unique weighted combinations of the features that each of a collection of regression trees applies to make a prediction.  This offers a relatively fast, minimally data hungry approach that can be used with many hardware platforms, making it accessible to niche communities for easy use and re-training for their specific algorithms and datasets.  We employ 25 trees and train by sampling one third of the predictive variables per decision split, sampling training examples with replacement, and requiring a minimum of five examples per tree leaf.  

\section{Experiments and Results}
\label{sec_ExperimentsAndResults}
We conduct studies to analyze the reliability of our prediction framework and its value for deciding when to target computers or humans to segment images.  

\textbf{Datasets.} We evaluate our methods on four datasets that represent three imaging modalities: Boston University Biomedical Image Library (BU-BIL:1-5)~\cite{GurariThSaIsPhPuSoWaZhWoBe15} includes 271 gray-scale images coming from three fluorescence microscopy image sets and two phase contrast microscopy image sets, Weizmann~\cite{AlpertGaBaBr07} consists of 100 grayscale images showing a variety of everyday objects, Interactive Image Segmentation~\cite{GulshanRoCrBlZi10} (IIS) contains 151 RGB images showing a variety of everyday objects, and MSRA10K~\cite{ChengMiHuToHu14} contains 10,000 RGB images showing a variety of everyday objects.  Each dataset contains human-drawn segmentations that serve as pixel-accurate ground truth segmentations.  

\begin{table}[b] \footnotesize  
         \centering
        \caption{Characterization of studied datasets to reveal the diversity of image content with respect to object area (\# pixels), centroid location (X Loc, Y Loc), shape (Sec.~\ref{sec_predictingSegQuality}; shape factor), and coverage in image ($\frac{\textrm{FG Area}}{\textrm{Image Area}}$) as well as image texture (FG Var, BG Var = variance of Laplacian values for object and background pixels respectively).}
        \begin{tabular}{| c | c | c | c | c | c | c | c | c | c | c |}
   \hline
      & \multicolumn{2}{|c|}{{\bf BU-BIL}} & \multicolumn{2}{|c|}{{\bf Weizmann}} & \multicolumn{2}{|c|}{{\bf IIS}} & \multicolumn{2}{|c|}{{\bf MSRA10K}} \\ \hline   
      & {\pmb{$\mu$}} & {\pmb{$\sigma$}} & {\pmb{$\mu$}} & {\pmb{$\sigma$}} & {\pmb{$\mu$}} & {\pmb{$\sigma$}} & {\pmb{$\mu$}} & {\pmb{$\sigma$}} \\ \hline \hline   
       \textbf{Area}  & 7,927 & 13,109 & 24,315 & 16,815 & 40,119 & 41,387 & 26,235 & 12,489 \\ \hline 
	   \textbf{X Loc} & 126 & 129 & 146 & 29 & 251 & 80 & 186 & 52 \\ \hline
       \textbf{Y Loc} &115 & 106 & 158 & 61 & 223 & 63 & 171 & 44  \\ \hline        
       \textbf{Shape} & 0.48 & 0.25 & 0.41 & 0.2 & 0.4 & 0.2 & 0.50 & 0.24 \\ \hline     
       \textbf{$\frac{\textrm{FG Area}}{\textrm{Image Area}}$} & 0.12 & 0.04 & 0.27 & 0.14 & 0.19 & 0.12 & 0.22 & 0.10 \\ \hline     
       \textbf{FG Var} & 54 & 51 & 1,663 & 1,271 & 2,227 & 1,909 & 1,292 & 1,244 \\ \hline       
       \textbf{BG Var} & 28 & 36 & 540 & 835 & 1,568 & 1,521 & 587 & 829 \\ \hline       
    \end{tabular}
    \label{table_datasetCharacterization}
\end{table}

Together, the four datasets exhibit large variability with respect to object and image properties (\textbf{Table~\ref{table_datasetCharacterization}}).  For example, the object size is over five times larger in IIS (i.e., 40,119 pixels) than in BU-BIL (i.e., 7,927 pixels).  The object consumes more than two times the area of the image in Weizmann (i.e., 0.27) than in BU-BIL (i.e., 0.12).  Moreover, there is rich diversity of object appearances within each dataset, as revealed by large {\pmb{$\sigma$}} values.  For example, there is a large Shape {\pmb{$\sigma$}} for all datasets.  Additionally, the variability of object texture (i.e., FG Var {\pmb{$\sigma$}}) is relatively large for all datasets.  Furthermore, our analysis suggests that image backgrounds can be complicated and/or cluttered (i.e., large BG Var {\pmb{$\mu$}} and {\pmb{$\sigma$}}).  The observed diversity of dataset characteristics is important to ensure our method is challenged to learn generic cues predictive of segmentation failure.  

\subsection{Quality Prediction for Algorithm Set}
\label{sec_predictRegionProposalQuality}
We first analyze the predictive power of our proposed framework to automatically estimate the quality of foreground object segmentations (Section~\ref{sec_predictingSegQuality}).  

\textbf{Baseline.} 
We compare our method to the \textit{CPMC~\cite{CarreiraSm10}} approach that also can predict the quality of any given object segmentation.  Specifically, it predicts a Jaccard score per segmentation.  This baseline stresses generality by learning statistics typical for real world objects.  The method learns to predict Jaccard scores on everyday images using a combination of shape and intensity-based features.  We use publicly-available code.  We do not compare against methods that return a relative ranking of proposal regions per image (e.g.,~\cite{EndresHo10}), because they are inadequate for ranking segmentations across a batch of images.  

\textbf{Evaluation Metrics.} 
We evaluate each prediction model using Pearson's correlation coefficient (CC) and mean absolute error (MAE).  CC indicates how strongly correlated predicted scores are to actual Jaccard scores for all foreground object segmentations evaluated.  Values range between +1 and -1 inclusive, with values further from 0 indicating stronger predictive power.  MAE is the average size of prediction errors, computed as the mean absolute difference between all predicted and actual Jaccard scores.  

\textbf{Ours: Cross-Dataset Generalization.} 
To minimize concerns that prediction successes may be due to over-fitting to the statistics of a particular dataset, we first evaluate how well our prediction models trained on three of the datasets perform on the fourth dataset\footnote{To afford similar contributions of each dataset, we randomly sample 2000 segmentations for the MSRA10K dataset.}.  We enrich our analysis by also examining the predictive performance of our models when trained and tested exclusively with the mask-based and intensity-based features respectively.  \textbf{Table~\ref{table_predScores}} shows our results when employing both mask-based and intensity-based features (row 2), intensity-based features alone which is the CNN features described in Section~\ref{sec_trainingFeatures} (row 3), and mask-based features alone which is all features described in Section~\ref{sec_trainingFeatures} except for the Intensity features (row 4).  For clarity in presenting the results, we only show results for the overall top-performing CNN feature, based on testing both with mask-based features and alone, from the three evaluated options: AlexNet~\cite{KrizhevskySuHi12}. 

\begin{table*}[t]
         \centering
        \caption{Comparison of CPMC~\cite{CarreiraSm10} with our method for predicting the Jaccard score indicating the quality of a foreground segmentation.  We report scores for our method learned with cross-set training (``C-Ours") and single-set training (``S-Ours") when using mask features alone (``-M"), intensity features alone (``-I"), as well as both mask and intensity features.  For intensity features, we consider three CNN options: AlexNet~\cite{KrizhevskySuHi12}, VGG~\cite{SimonyanZi14}, and ResNet~\cite{HeZhReSu16}.   We conduct experiments with four datasets.  Higher correlation coefficient (CC) scores and lower mean absolute error (MAE) scores are better.}
            \label{table_predScores}
        \begin{tabular}{| l | c | c | c | c | c | c | c | c | c | c | }
   \hline
      & \multicolumn{2}{|c|}{{\bf BU-BIL}} & \multicolumn{2}{|c|}{{\bf Weizmann}} & \multicolumn{2}{|c|}{{\bf IIS}} & \multicolumn{2}{|c|}{{\bf MSRA10K}}  \\ \hline   
      & \textbf{CC} & \textbf{MAE} & \textbf{CC} & \textbf{MAE} & \textbf{CC} & \textbf{MAE} & \textbf{CC} & \textbf{MAE} \\ \hline \hline   
       \textbf{C-CPMC~\cite{CarreiraSm10}}  & 0.18 & 0.30 & 0.27 & 0.30 & 0.23 & 0.32 & 0.61 & 0.25  \\ \hline
        \textbf{C-Ours} & 0.63 & 0.21 & 0.61 & 0.23 & 0.50 & 0.26 & \textbf{0.62} & \textbf{0.23}   \\ \hline   
        \textbf{C-Ours-I} & -0.10 & 0.33 & 0.40 & 0.27 & 0.41 & 0.29 & 0.49 & 0.26   \\ \hline  
        \textbf{C-Ours-M} & 0.66 & 0.19 & \textbf{0.65} & \textbf{0.21} & \textbf{0.56} & \textbf{0.23} & 0.59 & 0.22   \\ \hline    \hline 
       \textbf{S-Ours} & 0.87 & 0.10 & \textbf{0.85} & \textbf{0.15} & 0.83 & 0.16 & \textbf{0.86} & \textbf{0.12}   \\  \hline   
        \textbf{S-Ours-I} & 0.84 & 0.13 & 0.81 & 0.16  & \textbf{0.84} & \textbf{0.15} & 0.80 & 0.15  \\ \hline   
        \textbf{S-Ours-M} & \textbf{0.88} & \textbf{0.09} & 0.79 & 0.16 & 0.73 & 0.19 & 0.78 & 0.15  \\ \hline   
    \end{tabular}
\end{table*}

Overall, our approach performs well, as indicated by high CCs and low MAEs (\textbf{Table~\ref{table_predScores}}, row 2).  The significant improvement of our approach over CPMC on the biomedical images (e.g., row 2 versus 1 with CC of 0.63 versus 0.18) shows it is successful even when trained on completely disjoint datasets---what the system learned on everyday images (Weizmann, IIS, MSRA10K) can successfully be leveraged on biomedical images (BU-BIL).  This is possibly because algorithms tend to create binary masks that have consistent properties at various levels of success and failure severity, regardless of the dataset.  Our approach also yields improvements over CPMC on the everyday images (Weizmann, IIS, MSRA10K), highlighting a potential value of populating training data with images from different modalities (e.g., biomedical images) to promote learning generic algorithm behavior rather than over-fitting to properties of a particular dataset. 
   
We observe that most of the predictive power of our model stems from mask-based features.  Mask-based features (\textbf{Table~\ref{table_predScores}}, row 4) perform better than variants of our model that employ intensity-based features (rows 2--3) for all but one dataset, when comparing CC and MAE scores.  This reveals a plausible limitation that intensity features do not generalize as well for different objects observed in images captured with different image acquisition technologies (e.g., microscopes) and parameters (e.g., lighting).  This hypothesis is supported by the observation that relying on the off-the-shelf CNN feature alone yields negligible predictive power for the biomedical images (\textbf{Table~\ref{table_predScores}}, row 3).  Moreover, we hypothesize the intensity-based features leads to high MAE values because of an accumulation of errors from using a high dimensional feature space.  Our findings demonstrate the characteristics of segmentation errors are robustly and sufficiently learned from a small set of features describing the binary mask alone and remain relevant across domains.    
 
\textbf{Ours: Single-Dataset Analysis.} 
Having demonstrated the advantage of our approach over an existing state-of-art baseline (i.e., CPMC) for cross-dataset tests, we next examine the performance gain of using our model when evaluating our prediction framework for each dataset independently (i.e., Weizmann, IIS, BU-BIL, MSRA10K).  Specifically, we train and test using 10-fold cross-validation on each dataset separately.  We again enrich our analysis by examining the predictive performance of our models when also trained and tested exclusively with the mask-based and intensity-based features respectively.  \textbf{Table~\ref{table_predScores}} shows results from employing all features (row 5), intensity-based features alone (row 6), and mask-based features alone (row 7).

As to be expected, we consistently observe further performance improvements when focusing on individual datasets (rows 5--7) rather than across datasets (rows 2--4); e.g., CC improves from $\sim$0.60 across each dataset to $\sim$0.85 (\textbf{Table~\ref{table_predScores}}, row 2 versus 5) when using both mask-based and image-based features.  Interestingly, different sets of features are more predictive for different datasets.  For example, the most predictive features are mask-based for BU-BIL, intensity-based for IIS, and the combination of both for Weizmann and MSRA10K.  We hypothesize the predictive features stem from the distinct biases of individual datasets.  For example, we hypothesize mask-based features matter more for BU-BIL because the dataset has relatively stronger shape-based biases; e.g., as observed in \textbf{Table~\ref{table_datasetCharacterization}}, objects in BU-BIL exhibit relatively little variation in image coverage (i.e., {\pmb{$\sigma$}} = 0.04). 

\begin{figure}[t!] 
\centering
\includegraphics[width=0.8\textwidth]{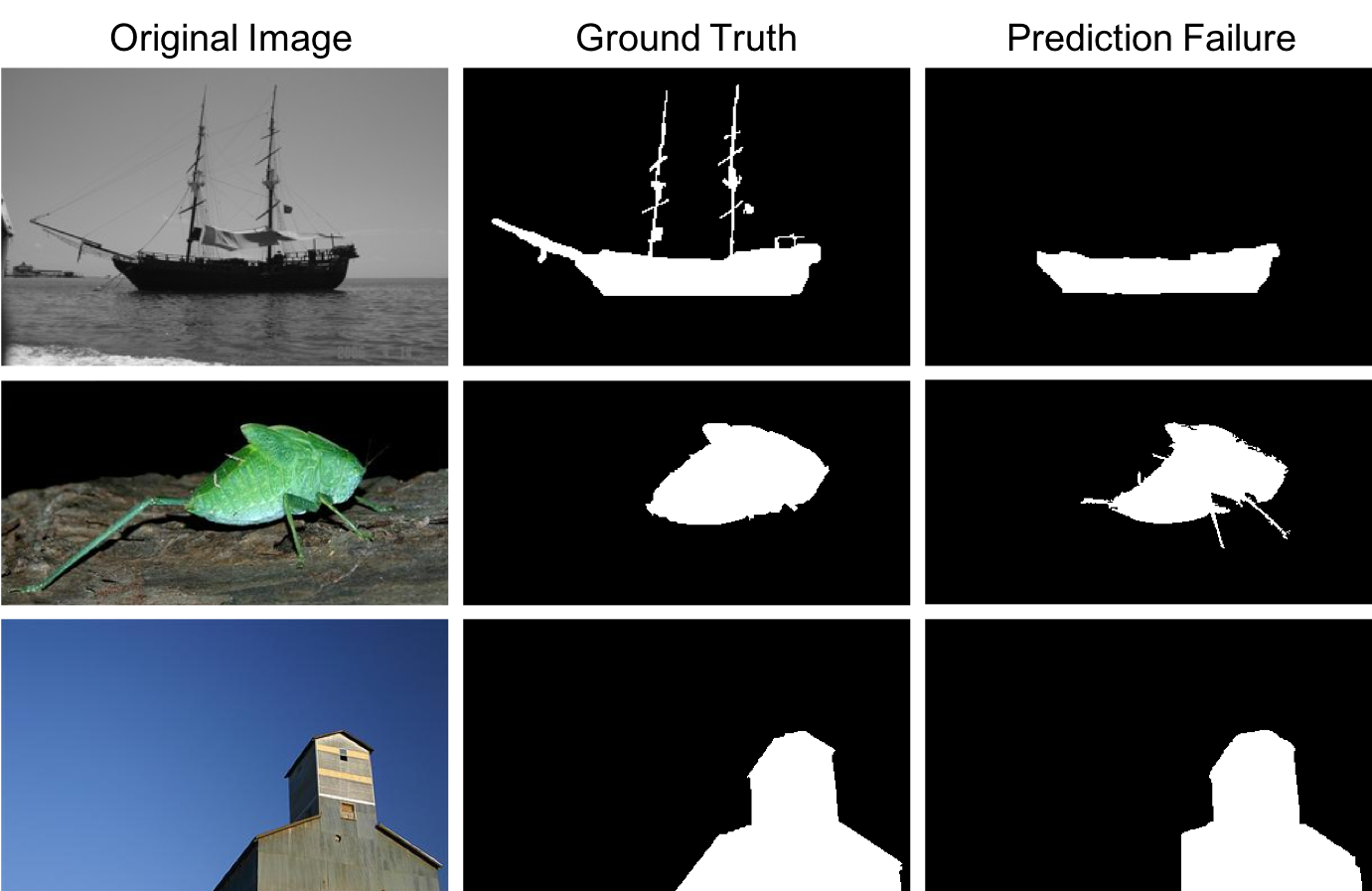}
\caption{Examples of images for which our prediction system makes inaccurate predictions.  Shown is the original image in the left column, ground truth foreground segmentation in the middle column, and an algorithm-drawn segmentation where our prediction module fails in the right column. }
\label{fig_predictionFailures}
\end{figure}

Our findings also highlight the impact of using different amounts of data for training.  Specifically, when comparing the performance of utilizing 10,000 images in MSRA10K versus $\sim$100-300 images in BU-BIL, Weizmann, and IIS for all feature combinations, we observe similar outcomes; i.e., CC scores range from 0.78 to 0.86 for MSRA10K which is slightly worse than the findings for BU-BIL and Weizmann and slightly better than the findings for IIS (\textbf{Table~\ref{table_predScores}}, rows 5--7).  This suggests that smaller datasets are sufficiently large for learning predictive cues that generalize\footnote{For the one dataset large enough to train a deep model, MSRA10K, we find that fine-tuning off-the-shelf CNNs (namely, AlexNet, VGG, and ResNet) yields similar or worse performance than the other models tested in our experiments, including those using the frozen CNN features without fine-tuning.   This suggests that the proposed features as is are well matched for the target task.}.  We also show qualitative results that illustrate some failure cases of our top-performing single-dataset prediction module, which employs mask-based feature vectors extracted from the AlexNet architecture (\textbf{Figure~\ref{fig_predictionFailures}}).  For all the examples, the predicted segmentation captures the main body of the object but receives low scores from the prediction module.  While the top and bottom examples are missing some object parts, the middle example appears to be penalized for capturing the highly-jagged edges on the boundary (excluded from the ground truth) despite that it still successfully captures the main body of the object.

Overall, our findings show it is possible to predict the quality of an image segmentation in absolute terms for a diversity of data spanning everyday and biomedical images.  As will be shown in the following sections, this capability offers exciting implications towards deciding which among multiple algorithms to choose to create the highest quality segmentations and deciding how to distribute effort between computers and humans to create high quality segmentations for a batch of images. 

\subsection{Analysis of Coarse Segmentation System} 
\label{sec_predictSrcForInteractiveSeg}
We now examine the value of our PTP framework for predicting whether to ``\textbf{P}ull \textbf{T}he \textbf{P}lug" on computers and solicit human effort for each image, when segmenting a batch of images with a given budget for human effort/time.  Our focus is on initializing segmentation tools.  The status quo is typically for humans to create a \textit{coarse object segmentation} input for every image.  However, computers also are sometimes employed to automatically position \emph{rectangles} based on the image dimensions~\cite{BernardFrThUn09,CasellesKiSa97,ChanVe01}.  Our system instead intelligently decides which among multiple automatic initialization methods is preferable for each image and then decides whether to involve humans instead (Section~\ref{sec_coarseSegSystem}).

\textbf{Implementation.}  For each image, our system deploys either (a) the algorithm-generated result from 14 options with the largest \textit{predicted} Jaccard score or (b) a human-drawn segmentation.  We leverage cross-dataset predictions (Section~\ref{sec_predictRegionProposalQuality}) from our top-performing mask-based predictors to estimate the quality of algorithm-drawn segmentations so that our method cannot inadvertently learn and exploit dataset-specific idiosyncrasies.  

\textbf{Baselines.}
We compare our method to the following hybrid human-computer methods for creating coarse segmentation inputs:
\\ \\
\noindent
- \textit{Perfect Predictor:}  For each image, this system deploys the algorithm-generated result from 14 options that has the largest \textit{actual} Jaccard score.  Human involvement is allocated to the images with lowest scores.  This predictor reveals the best initializations possible with our system.
\\ \\
\noindent
- \textit{Chance Predictor:} For each image, the system randomly deploys one algorithm-generated result from the 14 options.  Then, images for human involvement are randomly selected.  For a lay person who lacks specialized knowledge of which algorithms work well in their domain, this predictor illustrates the best (s)he can achieve today with the initialization options available in our system.
\\ \\
\noindent
- \textit{Rectangle~\cite{BernardFrThUn09,CasellesKiSa97,ChanVe01}:} This method illustrates the commonly-adopted automated method of positioning a bounding rectangle with respect to the image dimensions.  Following~\cite{ChanVe01}, we set the foreground region based on the image boundary.  We position the rectangle to occupy the image region after cropping 5\% of pixels from the minimum image dimension on all sides.  We randomly select images for human involvement.    
\\ \\
\noindent
- \textit{Linear~\cite{GurariJaBeGr16}:} This is a variant of our approach, as implemented in our prior experiments, that uses a linear regression model instead of the non-linear regression trees.  It predicts which of the original eight segmentation options to deploy as input.
\\ \\
\noindent
- \textit{No-Refinement}: This method illustrates the performance of our prediction system in the absence of refinement.  Specifically, the system relies on the input crowd-generated and algorithm-generated results as is, rather than the output of the refinement algorithms that modify these inputs.
\\ \\

\noindent
\textbf{Experimental Design.} To illustrate the versatility of our initialization system as a general-purpose approach for use with refinement segmentation tools, we integrate our initialization method and the baselines with three tools important in the computer vision and medical imaging communities - GrabCut~\cite{RotherKoBl04}, Chan Vese level sets~\cite{ChanVe01}, and Lankton level sets~\cite{LanktonTa08}.  GrabCut enforces color homogeneity and spatial proximity.  The Chan Vese level set method uses global image information to try to separate an image into two homogeneous intensity regions while enforcing smoothness of the object boundary.  The Lankton level set method uses local neighborhood statistics for each pixel to separate an object from the background so that there are two homogeneous intensity regions within a band containing the object boundary.  

We evaluate each system using all images in Weizmann, IIS, and BU-BIL as well as a random sample of 174 images from MSRA10K.  We evaluate with a subset of images from MSRA10K in order to make it comparable in size to the other datasets; 174 is the average number of images in Weizmann (i.e., 100), IIS (i.e., 151), and BU-BIL (i.e., 271).  We examine the performance of each initialization method for all 696 images coming from all four datasets as well as for each dataset independently. 

For human input, we collect segmentations from crowd workers on Amazon Mechanical Turk.  We use the same crowdsourcing system employed in prior work~\cite{JainGr13} to collect a coarse segmentation per image.

\begin{figure}[t!] 
\centering
\includegraphics[width=0.98\textwidth]{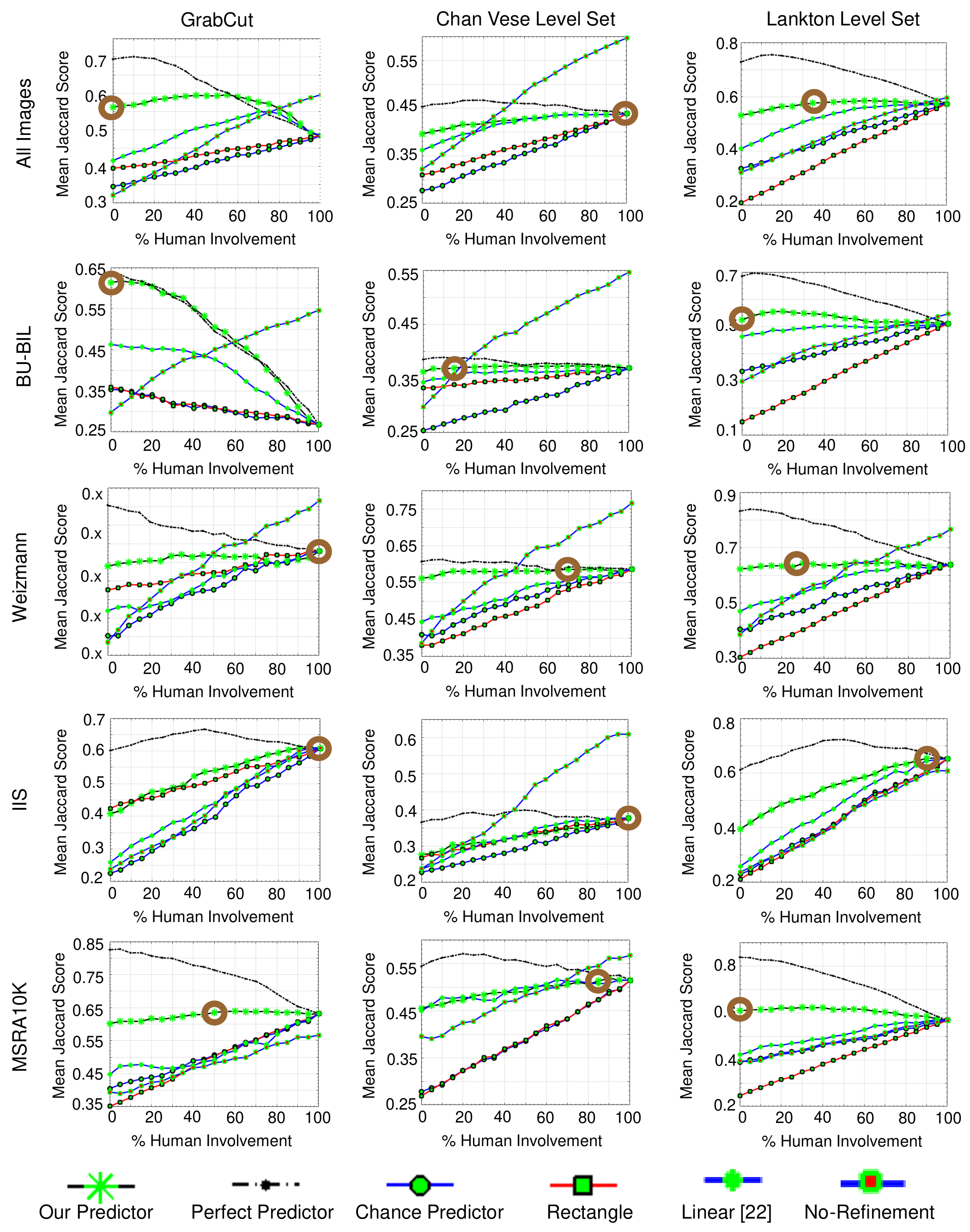}
\caption{We compare six methods for distributing varying levels of human involvement to create initializations for three different segmentation refinement tools (cols 1-3) across four datasets (row 1) and each dataset independently (rows 2--5).  Each plot shows the mean quality for the segmentations that resulted after the tools refined the initializations.  Brown circles identify when our predictor achieves quality comparable to relying exclusively on human input with the least human effort.  Compared to relying exclusively on human input, our approach eliminates 55\% of human effort with no loss to quality. }
\label{fig_batchModeAlgInitAnalysis}
\end{figure}

\begin{figure}[t!] 
\centering
\includegraphics[width=1\textwidth]{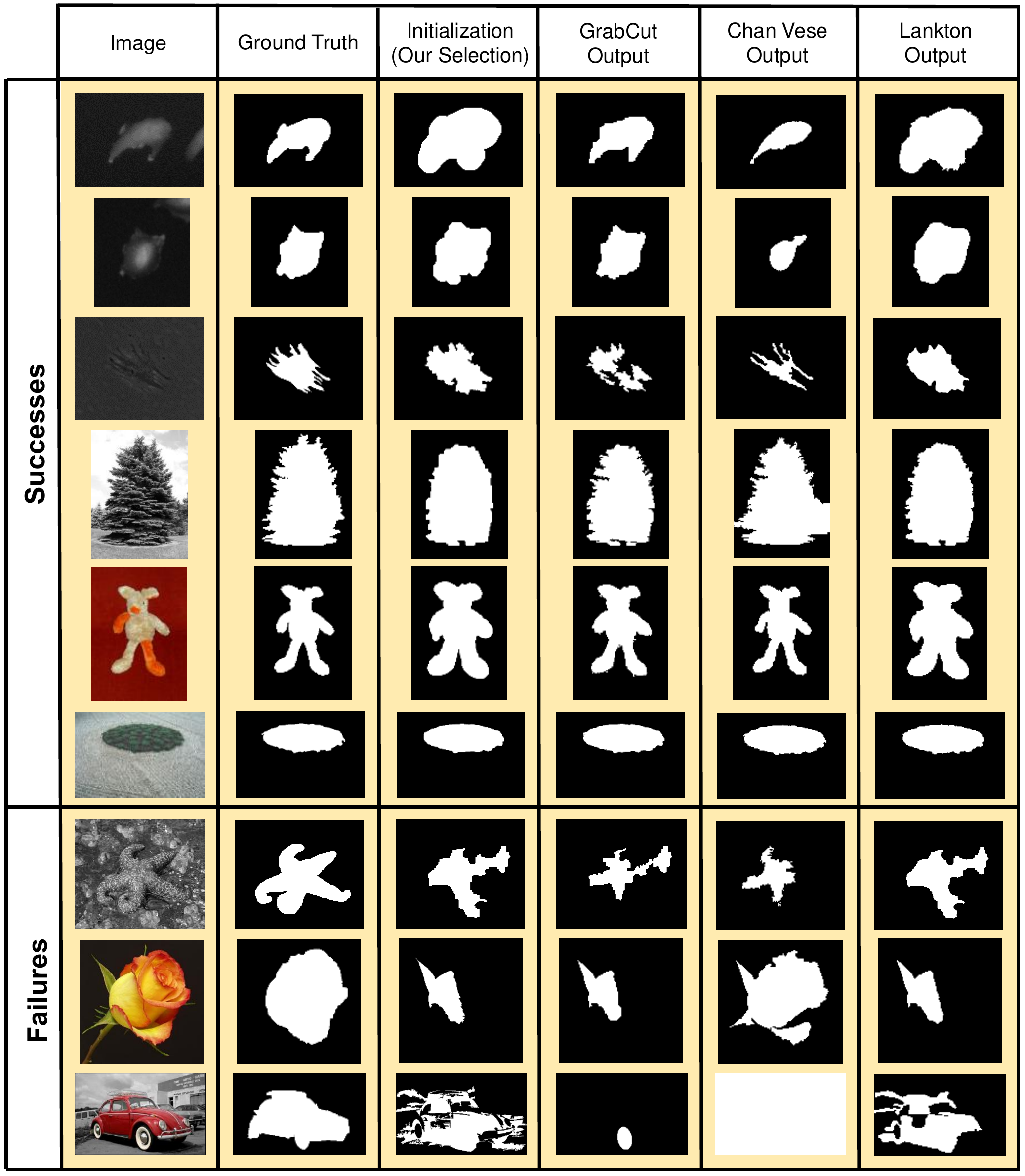}
\caption{Illustration of the quality of resulting segmentations created by three segmentation tools from the initial segmentation selected by our system from the 14 initialization options.  Sample results are shown for images coming from three imaging modalities: fluorescence microscopy (rows 1--2), phase contrast microscopy (row 3), and everyday images (rows 4--9).  These illustrate successes (rows 1--6) and failures (7--9) in creating high quality output segmentations.}
\label{fig_courseInitializationResults}
\end{figure} 

\subsubsection{Fully-Automated Initialization} 
\label{sec_fullyAutomatedInitialization}
 On average, our system takes 14.51 seconds to generate all candidate segmentations, 0.52 seconds to predict which result is the best, and 9.01 seconds for refinement (i.e., average of 1.33 seconds for GrabCut, 10.17 seconds for Chan Vese level set, and 15.53 seconds for Lankton level set).
 
For each segmentation tool, we compute the average segmentation quality resulting after the tool refines all initializations for all images.  For this analysis, the reader should focus on the leftmost points on the plots in \textbf{Figure~\ref{fig_batchModeAlgInitAnalysis}} only (i.e., 0\% human involvement).  

Predicting a best-suited automated input from 14 options produces coarse segmentation estimates that the segmentation tools can refine more successfully than all baselines (i.e., Chance Predictor, Rectangle~\cite{BernardFrThUn09,CasellesKiSa97,ChanVe01}, Linear~\cite{GurariJaBeGr16}, No-refinement).  For example, the resulting segmentation quality improves by 31 percentage points and 15 percentage points over the Rectangle baseline for the Lankton level set algorithm and GrabCut algorithm respectively (\textbf{Figure~\ref{fig_batchModeAlgInitAnalysis}}, ``All Images").  With respect to our Linear approach~\cite{GurariJaBeGr16}, we observe the greatest boost from our new approach on the Weizmann dataset; e.g,. 14 percentage points improvement for the Lankton level set algorithm.  The only exception where our proposed algorithm does not yield better results to the baselines is with the GrabCut algorithm on the IIS dataset; i.e., the Rectangle baseline performs better by approximately two percentage points.  We hypothesize this exception is because the images typically show more complex scenes and backgrounds, as suggested by the high pixel foreground and background variance in \textbf{Table~\ref{table_datasetCharacterization}}, which causes initializations to sometimes latch on to a region that does not contain the object of interest.  Overall, our findings highlight the value of intelligently predicting a best-suited algorithm per image from multiple options rather than relying on a single initialization approach.  

\begin{figure}[t!] 
\centering
\includegraphics[width=1\textwidth]{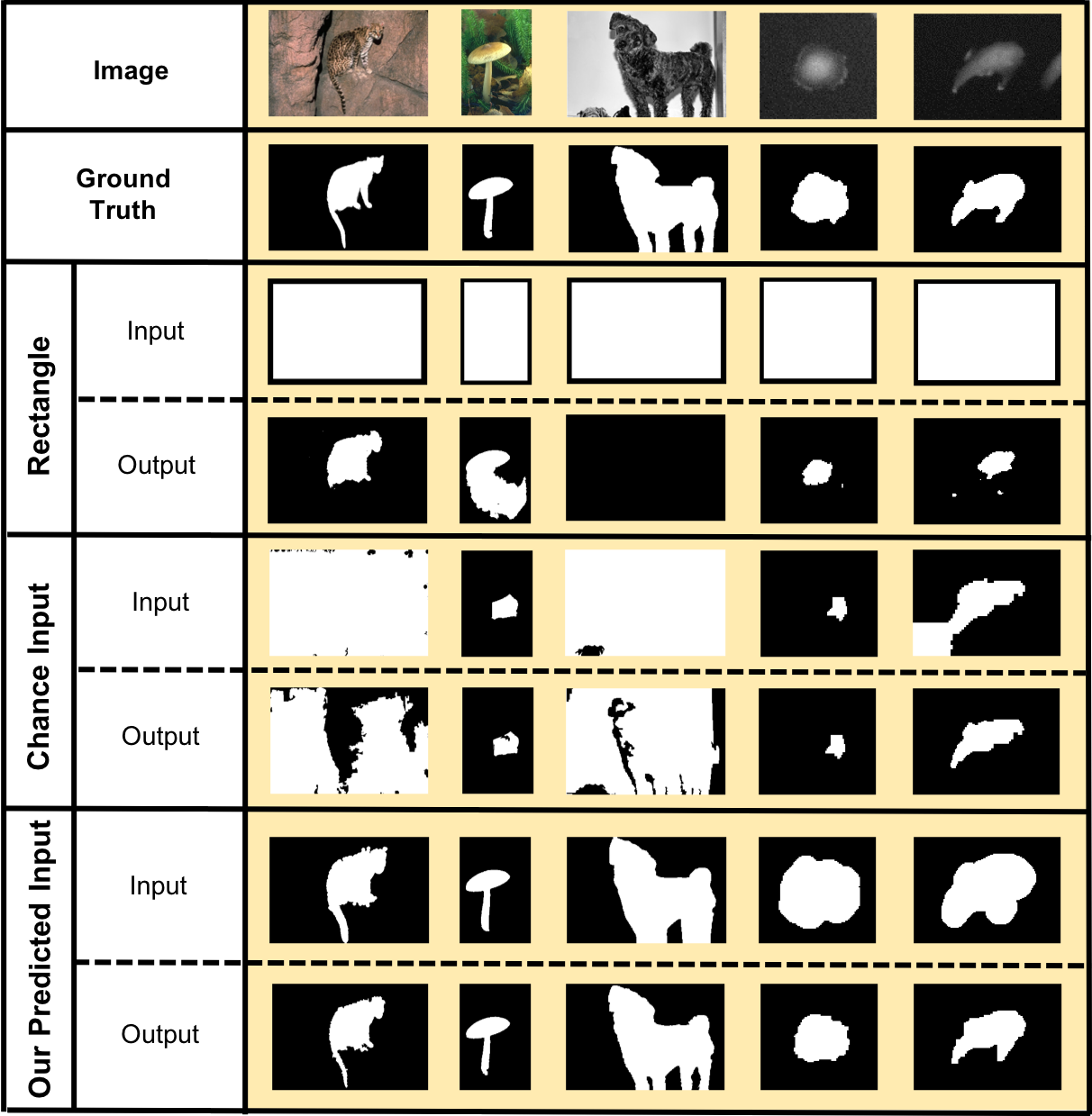}
\caption{Performance of the GrabCut algorithm when refining three different initialization approaches (``Input") into final (``Output") segmentations.}
\label{fig_exampleResultsForGrabCutsInitializedDifferently}
\end{figure}  

We show qualitative results that illustrate the versatility of our system to initialize the three different segmentation tools in \textbf{Figure~\ref{fig_courseInitializationResults}}.  As shown, given the same initialization, the three segmentation tools can produce very similar segmentations in some cases (e.g., plot of land in row 6) and dramatically different segmentations in other cases (e.g., spiculated cell shown in row 3).   We also observe that a segmentation tool can perform well when using a low quality initialization, as observed for the image of the cell (row 1, GrabCut algorithm).  These qualitative results illustrate our quantitative finding that applying a refinement algorithm typically yields considerable improvements over using the input as is (``No-Refinement"); e.g,. by over 20 percentage points for the Lankton level set algorithm (\textbf{Figure~\ref{fig_batchModeAlgInitAnalysis}}, ``All Images").  More generally, our system can often automatically produce sufficiently accurate initializations required by segmentation tools to produce segmentations that resemble the ground truth.  

We also show qualitative results that illustrate the failure cases of our system (\textbf{Figure~\ref{fig_courseInitializationResults}; rows 7--9}).  In some cases, none of the three segmentation tools perform well when initialized poorly, as observed for the image of the starfish (row 7) and flower (row 8).  Additionally, the refinement algorithms can perform poorly in refining an initialization, as observed for the image of the car which has a noisy initialization that roughly segments it out and refinements that do not restore the car (row 9).

We also show results comparing the output from the GrabCut algorithm initialized with our fully-automated segmentation initialization system as well as two baselines in \textbf{Figure~\ref{fig_exampleResultsForGrabCutsInitializedDifferently}}.  As observed, the quality of segmentation results is higher with our intelligent selection approach than arbitrarily chosen initial segmentation estimates (i.e., Rectangle, Chance).  Still, our approach does not necessarily ensure all fine-grained details of the object boundary are captured, as observed in columns 1 and 4.    

Our results highlight that applying the refinement algorithm to the top-predicted coarse segmentation leads to improvements over ``No-Refinement" for all three refinement algorithms: GrabCut, Chan Vese level set, and Lankton level set.  For example, we observe a 26 percentage improvement when applying the GrabCut algorithm rather than using the coarse segmentations alone (\textbf{Figure~\ref{fig_batchModeAlgInitAnalysis}}, ``All Images").  This demonstrates a benefit of applying refinement algorithms to clean up coarse segmentations.

\subsubsection{Reducing Human Initialization Effort}  
Thus far we have analyzed the impact of our method in a fully unsupervised setting, i.e., 0\% human involvement.  We next examine the impact of actively allocating human involvement to create \textit{coarse segmentation input} as a function of the budget of human effort available.  For each segmentation tool, we compute the average segmentation quality resulting after the tool refines the collection of chosen computer and human initializations for all images.  These results are also shown in \textbf{Figure~\ref{fig_batchModeAlgInitAnalysis}}; i.e., all values greater than 0\% human involvement.  

Our approach typically outperforms random decisions (i.e., Chance, Rectangle) and our Linear approach~\cite{GurariJaBeGr16} regarding how to distribute the initialization effort to humans and computers for all budget levels across all datasets.  Our approach also has the potential to outperform all three baselines for all segmentation tools by greater margins given improved prediction accuracy, as exemplified by the Perfect Predictor.  

In the more challenging setting of eliminating human effort without compromising segmentation quality, our system yields exciting results.  Specifically, our system achieves comparable quality to relying exclusively on crowdsourced input (i.e., 100\% human involvement) while using no human involvement for all images for GrabCut and human involvement for 35\% of images for Lankton level sets (\textbf{Figure~\ref{fig_batchModeAlgInitAnalysis}}; see brown circles on figures).  Our results reveal that different segmentation tools can tolerate different amounts of unreliable computer input without compromising the overall segmentation quality attained when relying exclusively on human input. 

Our findings also highlight what, if any, benefits arise from employing refinement algorithms versus using the input segmentations as is.  Overall, we observe that two of the three refinement algorithms yield considerable improvements for most budget levels across all datasets: GrabCut and Lankton level sets.  However, we found for these algorithms that it is beneficial to forego refinement when the available human budget can cover 80\% or more of the images; i.e., the No-Refinement results tend to outperform Our Predictor.  Our findings underscore the importance of selecting a good refinement algorithm and understanding its strengths in order to reap the benefit of our coarse segmentation initialization system.  
  
Given the scalability of crowdsourcing, we employed (non-trained) crowd workers to provide human annotations.  However, employing trained experts instead could lead to higher quality results from human initializations and so slightly different quality-human effort trade-offs.

\subsubsection{Peak Segmentation Quality}  
\label{sec_peakSegQuality}
Relying on a mix of human and computer efforts can outperform relying on either resource alone to create initial segmentations.  For example, peak accuracy for GrabCut with our initialization approach is achieved with 55\% human and 45\% computer involvement (\textbf{Figure~\ref{fig_batchModeAlgInitAnalysis}}, GrabCut on ``All Images").  There is a 12 percentage point improvement from relying on a mix of human and computer input over human input alone.  We attribute this finding to the tool's shrinking bias, which leads GrabCut to perform poorly when the initial boundary does not entirely subsume the true object region.  We believe this tendency is especially pronounced for human-drawn course segmentations for the biomedical images, as suggested by the algorithm consistently performing poorly when converting these to final segmentations (\textbf{Figure~\ref{fig_batchModeAlgInitAnalysis}}, GrabCut on ``BU-BIL"; 100\% human involvement).  In addition, we observe slight performance gains for the Lankton level set algorithm, with the tool fluctuating around a peak plateau value from 35\% to 100\% human involvement (\textbf{Figures~\ref{fig_batchModeAlgInitAnalysis}}, Lankton level set on ``All Images").  We attribute the latter performance fluctuations to slight differences when the tool expands and shrinks the human and algorithm initializations as needed to recover the desired boundaries.  More generally, our findings reveal that intelligently replacing human effort with computer effort can be desirable not only to save money and time, but to also collect higher quality segmentations.  

Our findings also demonstrate that the best possible performance across all benchmarked methods is obtained with the Perfect Predictor for two of the three refinement algorithms (GrabCut and Lankton level set) by relying on more computer effort than human effort.  As observed, the peak performance arises at $\sim$10\% human involvement for both algorithms.  A natural question is why does the performance increasingly fall with increasing amounts of human effort.  We hypothesize this fall is partially due to the relative lower quality of human-annotated coarse segmentations compared to what is possible with a fully automated approach.  Specifically, while the average quality for all human annotations is $\sim$58\% (\textbf{Figure~\ref{fig_batchModeAlgInitAnalysis}}, ``No-Refinement" for ``All Images"; 100\% human involvement), a fully-automated approach yields on average 71\% (i.e., 0\% involvement for Perfect Predictor for GrabCut and Lankton level set algorithms).  We also hypothesize the performance drop arises partially because of inadequate performance from the refinement algorithms, since refining reasonably high quality coarse segmentations can lead to worse performance than using the coarse segmentations as is for all three refinement algorithms; e.g., $\sim$60\% for ``No-Refinement" versus $\sim$45\% for all remaining methods for the GrabCut algorithm (\textbf{Figure~\ref{fig_batchModeAlgInitAnalysis}}; ``No-Refinement" versus all other methods for ``All Images"; 100\% human involvement).  

\subsection{Analysis of Fine-Grained Segmentation System} 
Lastly, we examine the value of our \emph{PTP} framework to predict when to pull the plug on computers and use human annotation instead to create fine-grained segmentations.  For this second task, given segmentations from algorithms, the system predicts which images humans should re-annotate in order to recover from failures (Section~\ref{sec_fineGrainedSystem}).  

\textbf{Implementation.}  Our system automatically feeds initializations from our fully-automated \emph{Coarse Segmentation} system to the GrabCut algorithm, the top-performing method found in Section~\ref{sec_fullyAutomatedInitialization}.  Quality estimates of the resulting segmentations are then predicted using our top-performing mask-based predictor (Section~\ref{sec_predictRegionProposalQuality}).  To avoid inadvertently learning and exploiting dataset-specific idiosyncrasies, we again employ the cross-dataset predictors.

\textbf{Baselines.} To our knowledge, no prior work addresses predicting when to enlist human versus computer effort for segmentation.  We compare our method to the following related methods for creating fine-grained segmentations:
\\ \\
\noindent
- \textit{Perfect Predictor:}  For each image, the system deploys the initial algorithm-generated result from 14 options that has the largest \textit{actual} Jaccard score, as done in Section~\ref{sec_predictSrcForInteractiveSeg}.  Then, the system ranks the resulting segmentations from the GrabCut algorithm based on the \textit{actual} Jaccard scores.  Human involvement is allocated to the images with lowest scores.  This predictor reveals the best results possible with our \textit{Fine-Grained Segmentation} system.
\\ \\
\noindent
- \textit{Chance Predictor:} For each image, the system deploys the commonly-adopted initialization of positioning a bounding rectangle with respect to the image dimensions (Section~\ref{sec_predictSrcForInteractiveSeg}, \emph{Rectangle}).  Then, images for human involvement are randomly selected.  This predictor illustrates the best a user can achieve today.
\\ \\
\noindent
- \textit{\textit{Jain \& Grauman}~\cite{JainGr13} (\emph{J \& G}) :} This method predicts how to best allocate a given budget of human time to annotate a batch of images.  In particular, it predicts whether to have humans draw a segmentation from scratch (54 seconds) versus supply a rectangle (7 seconds) or coarse segmentation (20 seconds) as input to GrabCut.  The system was trained on everyday images for GrabCut.  We use publicly-available code.  
\\ \\

\textbf{Experimental Design.} To represent images from the three imaging modalities with a similar number of images per modality, we conduct studies on all images from Weizmann, IIS, and BU-BIL.  Following prior work~\cite{JainGr13}, we budget 54 seconds for each segmentation a human creates from scratch.  We examine the impact of actively allocating human effort using a budgeted approach, ranging from no human involvement (0 minutes) to getting all images from the three datasets manually annotated (470 minutes).  We compute the average segmentation quality resulting for all chosen human-drawn and computer-drawn segmentations at each allotted time budget.

For human input, we collect segmentations from online crowd workers.  We measure the quality as the Jaccard similarity of each crowdsourced segmentation to the ground truth.  

\begin{figure}[t!]
\centering
\begin{subfigure}{0.5\textwidth}
  \includegraphics[width=1\linewidth]{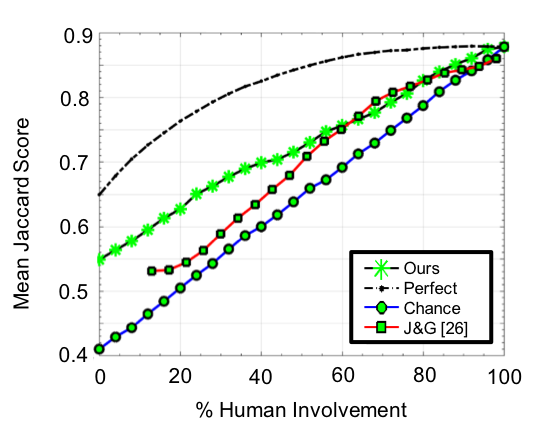}
    \caption{}
    \vspace{2em}
\end{subfigure}
\begin{subfigure}{0.49\textwidth}
  \centering
  \includegraphics[width=1\linewidth]{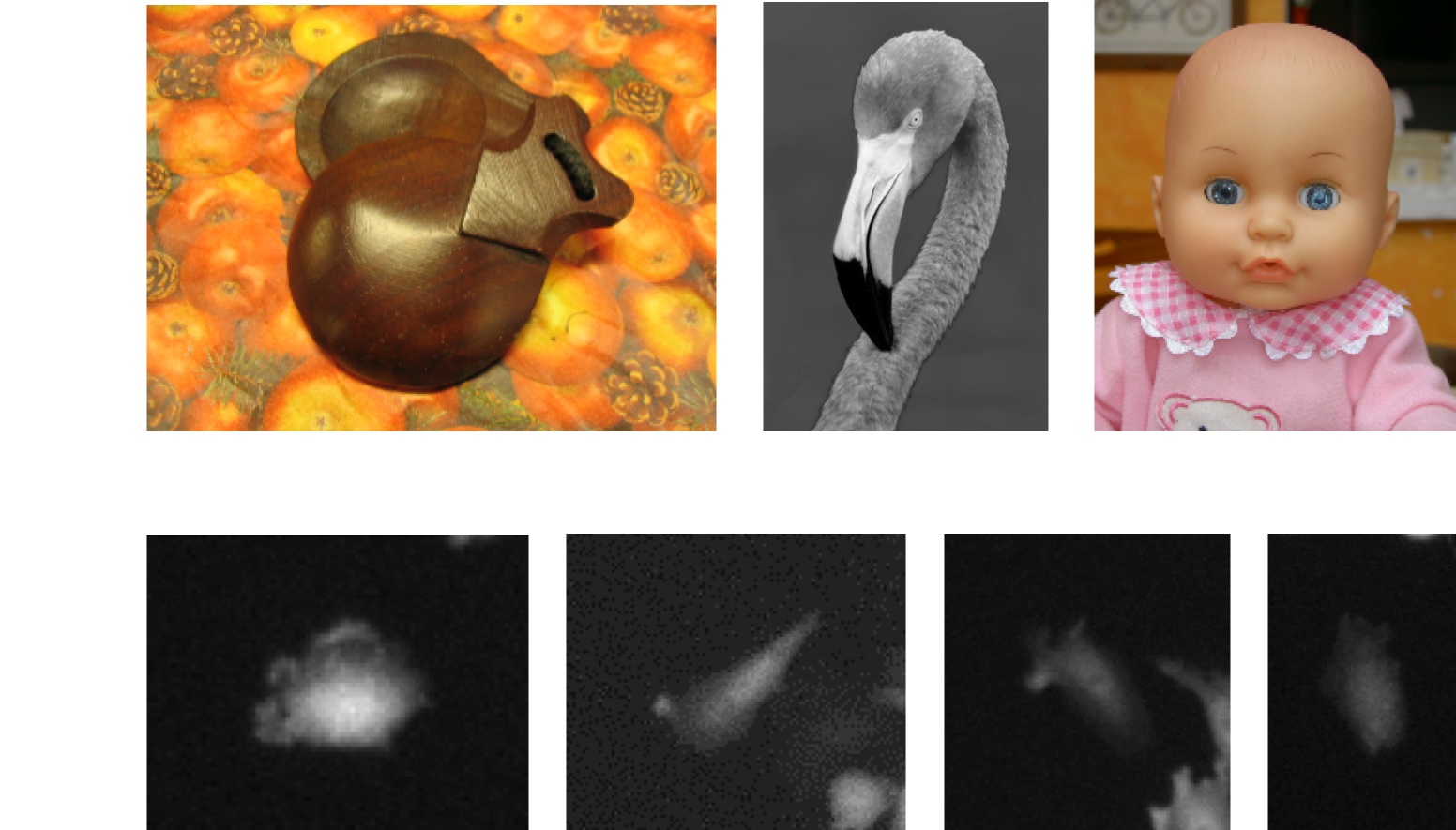}
  \caption{}
\end{subfigure}
\vspace{-1.5em}
\caption{Predicting when to replace segmentations created by a semi-automatic segmentation tool with segmentations created by (\textbf{a}) online crowd workers for 522 images representing three imaging modalities.  Our system typically achieves state-of-art accuracy (\emph{J \& G} method~\cite{JainGr13}) while saving up to 100 minutes of human effort (i.e., time difference between curves in the human budget range of 0\% to 20\% human involvement). (\textbf{b}) Examples of images which computers segment more similarly to experts than crowd workers.  As intended, our system often avoids involving crowd workers for these images.}
\label{fig_tightBoundaryPredictions}
\end{figure}

\textbf{Results.} Our system typically outperforms the related state of art \emph{J \& G} interactive method~\cite{JainGr13} for a wide range of budgets (\textbf{Figure~\ref{fig_tightBoundaryPredictions}a}).  The benefit of our approach is greatest in the range of 0-40\% human involvement, typically eliminating 45-100 minutes of human annotation effort to achieve segmentation quality comparable the \emph{J \& G} interactive method~\cite{JainGr13}.  A further advantage of our approach is that, unlike the \textit{J \& G}~\cite{JainGr13} system, our system works even when human involvement is not available for every image.  Specifically, as observed in \textbf{Figure~\ref{fig_tightBoundaryPredictions}a}, the \emph{J \& G} method~\cite{JainGr13} only becomes relevant at the budget level that supports human-created bounding boxes for all images (i.e., approximately at 12\% human involvement).  Our findings highlight the value of our system to save human effort.

Our findings also highlight the importance of a strong predictor for our system.  For example, with no human involvement, our proposed approach could improve a further 10 percentage points to achieve the performance of the \emph{Perfect Predictor} (\textbf{Figure~\ref{fig_tightBoundaryPredictions}a}).  Furthermore, our system would yield comparable quality to relying exclusively on crowdsourced workers while eliminating 16 percentage points of human effort, given a perfect predictor (\textbf{Figure~\ref{fig_tightBoundaryPredictions}a}, \emph{Perfect Predictor}).  While there are clear benefits from our approach, a valuable area for future work is to further improve the predictor.  For example, while our quality prediction system is currently designed to be agnostic to the refinement algorithm, it could instead be retrained using masks generated by the refinement algorithm towards improving its performance. 

Our findings also reveal that relying on a mix of human and computer effort can outperform methods that always assume human involvement.  In particular, for the last 55 images assigned to receive human annotations (i.e., images with highest predicted algorithm scores), the system appropriately chooses computer-drawn segmentations over human-drawn segmentations for 16\% of images.  For those images, computers create segmentations more similar to the ground truth than crowd workers (i.e., higher Jaccard scores).  Example images where algorithms segment better than the crowd are shown in \textbf{Figure~\ref{fig_tightBoundaryPredictions}b}.   

\section{Conclusions}
We proposed two novel tasks for intelligently distributing segmentation effort between computers and humans.  Both tasks rely on our proposed prediction module that predicts the quality of candidate segmentations from three diverse modalities (i.e., visible, phase contrast microscopy, fluorescence microscopy).  For the first task of creating initializations that segmentation tools refine, our proposed system eliminated the need for crowdsourced human annotation effort for an average of 55\% of images while preserving the resulting segmentation quality achieved when relying exclusively on human input.  For the second task of creating high quality segmentation results, our proposed system consistently preserved the resulting segmentation quality from a state of art interactive segmentation tool while regularly eliminating human annotation time.  Our work can relieve lay people from requiring domain expertise to identify which segmentation algorithm to use by automatically identifying which from numerous popular algorithms is best.  Moreover, it guides end users to direct their limited time to where their efforts will be of most value.
  
Valuable future research would include generalizing this work by designing a larger-scale system that supports more algorithms and image sets.  Towards this aim, next steps include creating a centralized, online repository of segmentation algorithms to which anyone can contribute and identifying the ideal, complementary subset of algorithms to use in order to avoid the computational overhead of applying all segmentation algorithms to each image.  Next steps also include generalizing the idea of automatically soliciting human input when algorithms fail for other tasks such spatio-temporal tracking of objects in videos.  Key issues to address include how to avoid error drift through the 3D image stack and what amount of 3D context should be presented to humans to support their video annotation efforts.

\section*{Acknowledgments}
\noindent
The authors thank the anonymous crowd workers for participating in our experiments.  This work is supported in part by National Science Foundation funding to DG (IIS-1755593), gift from Adobe to DG, National Science Foundation funding to MG (IIS-1421943), AWS Machine Learning Research Award to KG, IBM Faculty Award to KG, IBM Open Collaborative Research Award to KG, and a gift from Qualcomm to KG. This is a post-peer-review, pre-copyedit version of an article published in \textit{International Journal of Computer Vision}. The final authenticated version is available online at: http://dx.doi.org/10.1007/s11263-019-01172-6.

\end{document}